\documentclass[sigconf, authorversion]{acmart}

\usepackage{todonotes}
\usepackage{subfig}
\usepackage{xcolor}
\usepackage{diagbox}
\usepackage{acro}
\DeclareAcronym{AC}{short=AC,long=Alternating Current}
\DeclareAcronym{ADL}{short=ADL,long=Activities of Daily Living}
\DeclareAcronym{NILM}{short=NILM,long=Non-Intrusive Load Monitoring}
\DeclareAcronym{MAE}{short=MAE,long=Mean Average Error}
\DeclareAcronym{RMSE}{short=RMSE,long=Root Mean Square Error}
\DeclareAcronym{NILMTK}{short=NILMTK,long=Non-Intrusive Load Monitoring Toolkit}
\DeclareAcronym{CO}{short=CO,long=Combinatorial Optimization}
\DeclareAcronym{DSC}{short=DSC,long=Discriminative Sparse Coding}
\DeclareAcronym{FHMM}{short=FHMM,long=Exact Factorial Hidden Markov Model}
\DeclareAcronym{Hart85}{short=Hart85,long=Edge Detection}
\DeclareAcronym{DAE}{short=DAE,long=Denoising Autoencoder}
\DeclareAcronym{RNN}{short=RNN,long=Recurrent Neural Network}
\DeclareAcronym{S2S}{short=S2S,long=Sequence-to-Sequence Optimization}
\DeclareAcronym{S2P}{short=S2P,long=Sequence-to-Point Optimization}

\AtBeginDocument{%
  \providecommand\BibTeX{{%
    \normalfont B\kern-0.5em{\scshape i\kern-0.25em b}\kern-0.8em\TeX}}}

\begin{document}


\title{Exploring Bayesian Surprise to Prevent Overfitting and to Predict Model Performance in Non-Intrusive Load Monitoring}



\author{Richard Jones}
\affiliation{%
  \institution{School of Engineering Science\\Simon Fraser University, Canada}
}
\email{rtj4@sfu.ca}

\author{Christoph Klemenjak}
\orcid{0000-0002-0113-6351}
\affiliation{%
  \institution{Institute of Networked and Embedded Systems\\University of Klagenfurt, Austria}
}
\email{klemenjak@ieee.org}

\author{Stephen Makonin}
\affiliation{%
  \institution{School of Engineering Science\\Simon Fraser University, Canada}
}
\email{smakonin@sfu.ca}

\author{Ivan V. Baji\'c}
\affiliation{%
  \institution{School of Engineering Science\\Simon Fraser University, Canada}
}
\email{ibajic@ensc.sfu.ca}

\renewcommand{\shortauthors}{Jones et al.}

\begin{abstract}
Non-Intrusive Load Monitoring (NILM) is a field of research focused on segregating constituent electrical loads in a system based only on their aggregated signal. Significant computational resources and research time are spent training models, often using as much data as possible, perhaps driven by the preconception that more data equates to more accurate models and better performing algorithms. When has enough prior training been done? When has a NILM algorithm encountered new, unseen data? This work applies the notion of Bayesian surprise to answer these questions which are important for both supervised and unsupervised algorithms. We quantify the degree of surprise between the predictive distribution (termed postdictive surprise), as well as the transitional probabilities (termed transitional surprise), before and after a window of observations. We compare the performance of several benchmark NILM algorithms supported by NILMTK, in order to establish a useful threshold on the two combined measures of surprise. We validate the use of transitional surprise by exploring the performance of a popular Hidden Markov Model as a function of surprise threshold. Finally, we explore the use of a surprise threshold as a regularization technique to avoid overfitting in cross-dataset performance. Although the generality of the specific surprise threshold discussed herein may be suspect without further testing, this work provides clear evidence that a point of diminishing returns of model performance with respect to dataset size exists. This has implications for future model development, dataset acquisition, as well as aiding in model flexibility during deployment.
\end{abstract}




\maketitle



\section{Introduction}

Non-Intrusive Load Monitoring (NILM), often referred to as load disaggregation, dates back to the seminal work presented in \cite{hart1985prototype}. In a nutshell, NILM describes the problem of identifying present electrical appliances within a time series consisting of a sequence of (power) measurements taken at a central point in the distribution grid of a building. As can be obtained from a recently-published review \cite{gopinath2020energy}, the number of NILM techniques relying on machine learning approaches, especially Deep Learning, has significantly increased during the past years. Compared to traditional NILM techniques, Deep Learning methods require considerably larger amounts of training data. Motivated by this, research groups have invested big efforts in collecting and publishing energy datasets. Energy datasets are the outcome of measurement campaigns in one or several buildings with the aim to collect energy consumption data at both the aggregate and load/appliance levels. \cite{pereira2018performance}.

In recent years, more and more energy datasets have emerged (e.g., \cite{makonin_ampds2, makonin2018rae, klemenjak2020synthetic, refit} to name a small few), which can vary considerably in terms of complexity, methodology, appliance characteristics and usage patterns, setting, etc. (e.g., see \cite{klemenjak2020towards, pereira2018performance}). With some datasets spanning several years of collection, considerable time and computational resources are spent in training new models. Newer approaches to NILM increasingly adopt deep learning methods (e.g., \cite{krystalakos18windowgru, harell2019wavenilm}), which can involve millions of tunable parameters, not to mention the often arduous process of hyperparameter tuning. It stands to reason, then, that effectively isolating the most important segments of a dataset relative to a model could improve time-to-deployment as well potentially regularize against overfitting. A common technique to truncate training time is to monitor the model's loss metric over a validation partition of the dataset. However, the entire available training set is used in an epoch before evaluation on the validation set is made. Given the wide variation in dataset complexity, arbitrarily training on a subset of the available data runs the significant risk of missing important relationships between appliance modes or even missing appliances modes entirely.

In an online setting, a common approach for disaggregation is to deploy generalized models that are subsequently specialized to a given house by an additional round of training \cite{zoha}. In these cases, appliance-level performance metrics are unavailable, and optimization of a model is instead left to crude estimates of performance such as internal consistency between proposed appliance profiles or extracted features, fraction of the total energy assigned, convergence of model parameters to specific values, etc. As a result, it can be difficult to know how much data is necessary to re-train generalized models. In a real use-case, consumers need to know when a NILM solution is accurate enough to be trusted. Additionally, appliances in a modern home can change abruptly. The addition, removal, or replacement of appliances in a home can quickly render inflexible models obsolete. To ensure the longevity of NILM solutions in residential homes, some measure of the novelty of incoming data is needed. If data can be recognized as even potentially useful in updating an existing model, these issues can be addressed.

The concept of novelty in incoming data has a model-specific dimension, in that different models may learn different features of the data. Clearly, data exhibiting novel features relative to those the model has already learned would qualify as novel or ``surprising''. Generalizing this notion of novelty is difficult and not amenable to a one-size-fits-all approach. However, there is also a way in which data can be intrinsically surprising, in the sense that specific appliance modes can be activated for the first time or exhibit abnormal behaviour. Moreover, appliances such as dish washers or clothes washers/dryers are multi-sequence machines with many user-operated programs. Data exposing new relationships between previously observed appliance modes may also qualify as intrinsically surprising. We approach both of these data-specific notions of novelty through the framework of Bayesian surprise.

The remainder of this paper is structured as follows: Section \ref{sec:related} gives a brief overview of the motivations behind Bayesian surprise and some of the previous work in the area. Section \ref{sec:nilm} relates these concepts to NILM by modeling appliance activations in a non-parametric Gaussian mixture model and introducing \textit{postdictive surprise}. Additionally, we introduce the concept of \textit{transitional surprise} by simply modeling the relationships between appliance states in a Markovian sense. Section \ref{sec:exp} shows some preliminary results, highlighting
\begin{enumerate}
    \item the diminishing returns of increased amounts of similar data,
    \item the potential ``model-agnostic'' regularization effect of training data truncation,
    \item and the usefulness of transitional surprise to (crudely) approximate system dynamics.
\end{enumerate}
Finally, Section \ref{sec:conclusions} provides some insights into the conducted experiments and some suggestions for further development of the concept of surprise in NILM.


\section{Related Work} \label{sec:related}

Literature defines surprise as the result of a discrepancy between expectation and observation, where expectation stems from experience gained through observation \cite{barto2013novelty}. As concerns surprise in the Bayesian framework, several techniques of Bayesian surprise measures have been proposed by related work.
In \cite{bayarri2003bayesian}, several measures of surprise are derived for outlier detection in normal models. On the basis of comparative studies, the authors recommend partial posterior predictive p-value and plug-in measures.

With regard to sequential (Bayesian) learning, Itti and Baldi \cite{itti2006bayesian, baldi2010bits} define \emph{Bayesian surprise} to be a measure of dissimilarity to assess the effect of data $D$ on the belief distributions of an observer. This means that Bayesian surprise can be understood as the distance (i.e. dissimilarity) between the prior distribution $P(M)$ and the posterior distribution $P(M|D)$ over a set $\mathcal{M}$ of possible models:

\begin{equation}
    \forall M \in \mathcal{M}, \quad P(M | D)=\frac{P(D | M)}{P(D)} P(M)
\end{equation}
\begin{equation}
S(D, \mathcal{M})= d[P(M|D), P(M)]
\end{equation}
where the relative entropy, or Kullback-Leibler (KL) divergence, is suggested to serve as distance measure $d$ in the initial proposal of \cite{baldi2002computational}. Instead of KL divergence, Jensen-Shannon and Cauchy-Schwarz can be used as well to compute Bayesian surprise, as done in \cite{hasanbelliu2012online}.

It can be observed that Itti and Baldi's interpretation of Bayesian Surprise has found application in various forms: de-biasing of thematic maps in \cite{correll2016surprise}, automatic detection of landmarks in computer vision \cite{ranganathan2009bayesian}, detection of salient acoustic events \cite{schauerte2013wow}, identification of calcifications in mammogram images \cite{domingues2014using}, and to determine suitable thresholds for extreme value models \cite{lee2015bayesian}.

In \cite{kolossa2015}, Bayesian updating of an agent's beliefs was grouped into two general categories. First, Bayesian surprise is the term given to the change in beliefs over latent variables, i.e., the divergence between the prior and posterior over unobservable quantities inferred through observations. Second, \textit{postdictive} surprise refers to the divergence between the prior and posterior predictive distributions, quantifying the surprise over observable quantities. In \cite{fancysurprise}, the concept of confidence-corrected surprise is developed, in which the degree of commitment to a particular generative model influences the extent to which observations update an agent's beliefs. However, given that the intent of the present work is to develop a ``model-agnostic'' formulation for NILM datasets, surprise in the present work is restricted to a fixed model (i.e., $\lvert \mathcal{M} \rvert = 1$).


An application of special interest to NILM turns out to be avoiding overfitting of algorithms during training, which is a common and unwanted effect when striving for accurate load disaggregators (with the aim to train good estimators). In particular, neural networks are prone to suffer from overfitting on a domain, especially when training is performed for too many iterations or with too little data \cite{incecco2020transfer}. Countermeasures for the overfitting problem have been developed and successfully been applied to NILM such as the early-stopping criterion, used in \cite{gomes2020pb, fagiani2019non}, dropout as in \cite{kim2017nonintrusive}, as well as sparsity or other norm constraints as in deep sparse coding~\cite{DSC} and related methods. Respectively, these approaches restrict the number of epochs of training based on the behaviour of the validation loss, prevent a random subset of parameters from being updated, or modify the loss itself to constrain the local minima to certain regions of the parameter space. The only method directly relating to the data itself is early-stopping, but it is model-specific in that it requires evaluation of the model trained over all available data. In order to have a \textit{data-centric} overfitting countermeasure that is applicable to all NILM techniques, it must be determinable without reference to the particular model being trained.

In \cite{hasanbelliu2012online}, the authors propose a Bayesian surprise metric based on the Cauchy-Schwarz divergence to differentiate between useful information and redundant observations during online learning of mixtures of Gaussians. The main motivation behind this measure is to prevent outliers from significantly changing the model parameters as well as restrict redundant samples from over-specifying component parameters, which would lead to overfitting. In the context of online learning, our work can be considered somewhat of an extension of \cite{hasanbelliu2012online} to non-parametric methods, rather than storing outliers and instantiating new components based on Gaussian Mean Shifting. However, the main focus of the present work is to use GMMs to explore the point at which the data is no longer surprising with respect to improving the performance of any model. By contrast, \cite{hasanbelliu2012online} uses the concept of Bayesian surprise within a GMM to optimize its own clustering performance.


\section{Surprise Methodology for NILM}
\label{sec:nilm}

A natural approach to characterize the novelty of incoming data is to examine the change in the signal and compare it to the changes so far observed. In other words, clustering on the first-differences of the signal permits an intuitive notion of surprising data: appliance events not yet seen. Following the basic appliance characterizations in \cite{hart1985prototype}, simple ON-OFF or multi-state appliances can have their initial activations modelled as Gaussian around some mean value.

However, transient characteristics of appliances, such as the consumption spike at the start of a fridge's condenser cycle,
can result in a highly varying activation value. Moreover, the consistency of these initial activations are dependent on sampling frequency. We consequently preprocess the data using a fast, steady-state block-filter developed in \cite{jones}. This filter imputes the mean value between change-points identified using an adaptive threshold on the raw power and first-differences in the signal. This steady-state power for individual appliance states is far more amenable to Gaussian modelling given its improved consistency. An example of the filter output and the corresponding raw aggregate data is shown in figure \ref{fig:filter}.

\begin{figure*}[ht]
\centering
    \subfloat{\centering
      \input{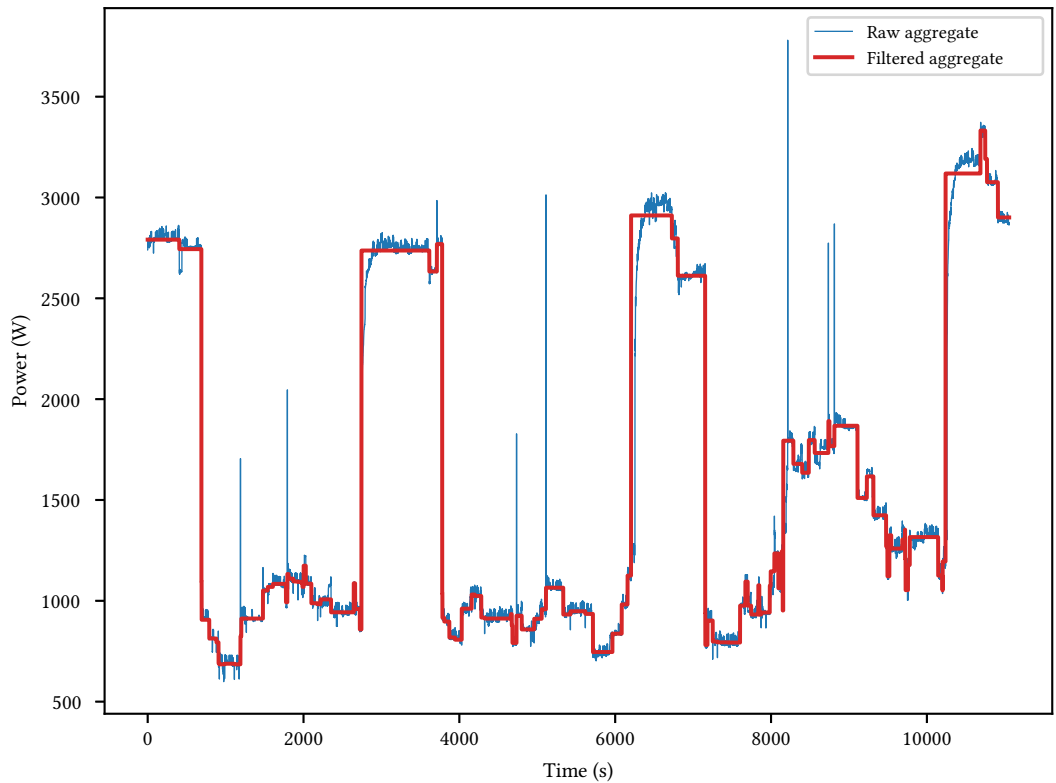}\label{fig:filter}
    }
    \caption{Example of Steady-state Block-filter output, from \cite{jones}}
    \label{fig:filter}
\end{figure*}

In a typical Gaussian mixture model with $K$ components, the likelihood is written as

\begin{equation}
    \label{eq:GMMlikelihood}
    p(x\lvert\theta_1,...,\theta_K) = \sum_{k=1}^K \pi_k \mathcal{N}\big(\mu_k,\Sigma_k\big),
\end{equation}
\noindent
where $\theta_k = \{\mathbf{\mu_k},\Sigma_k,\pi_k\}$ parameterizes component $k$ by its mean vector $\mathbf{\mu_k}$, its mixing proportion $\pi_k$ (where $0\leq \pi_k \leq 1$ and $\sum_k \pi_k = 1$), and its covariance matrix $\Sigma_k$.

In the Bayesian context, prior distributions are placed on each component's parameters, which in turn are parameterized by a set of hyperparameters shared across components. For the sake of inferential tractability, these priors are typically conjugate to their likelihoods. In the general case, component means and covariances are unknown, requiring a normal-inverse-Wishart joint prior, described for each component by

\begin{gather}
    \label{eq:NIW}
    \Sigma \stackrel{}{\sim} \text{IW}(\nu,\Delta) \nonumber\\
    \mathbf{\mu}\lvert\Sigma \stackrel{}{\sim} \mathcal{N}(\phi,\Sigma/\kappa),
\end{gather}
\noindent
Here, \text{IW} is the inverse-Wishart distribution with covariance/scale matrix $\Delta$ and degrees of freedom $\nu$. Similarly, the conditionally normal prior on the component means is parameterized by a base mean, $\phi$, and covariance scaled by another hyperparameter, $\kappa$.

The mixing proportions for each component are typically given a Dirichlet conjugate prior with hyperparameter $\alpha$:
\begin{equation}
    \label{eq:Dir}
    \mathbf{\pi}\lvert \alpha \stackrel{}{\sim} \textit{Dir}(\alpha_1,\alpha_2,...,\alpha_K),
\end{equation}

\noindent
where the $\alpha_i$'s are the ``pseudo-count'' prior observations of the $i^{th}$ component. Typically the prior is symmetric such that $\alpha_1 = \alpha_2 =...=\alpha_K = \alpha$.


This construction allows the parameters and weights of the $k$ Gaussian components to be sampled according to the data, often in Markov Chain Monte Carlo methods such as Gibbs sampling. Despite the inherent flexibility, GMM's are a parametric method, i.e., one of fixed dimensionality. Shifts in component weights when observing new data may be surprising, but this is a more gradual shift, and the predictive distribution will converge to a relatively stationary distribution that accounts for the prevalence of each component. Instead, the intuitively surprising aspect of new data is the instantiation of a new component/appliance state. This requires an extension of Gaussian mixtures into nonparametric methods, which we briefly overview.

In order to achieve an unbounded set of mixing components and their respective mixing proportions, we introduce the Dirichlet Process (DP). The DP is a stochastic process that generates random probability measures which follow a Dirichlet distribution for every finite partition of some measurable space~\cite{ferguson1973}. It is uniquely defined by a base measure on the measurable space and a concentration parameter, similar to the finite-dimensional Dirichlet distribution. The more intuitive ``stick-breaking'' picture of the DP was provided by \cite{stick}, which naturally motivates the use of DPs in mixture models as a nonparametric prior. In the stick-breaking procedure, the infinite sequence of mixing proportions are generated by drawing from a GEM distribution, described by
\begin{gather}
    \label{eq:GEM}
    \nu_i\lvert \alpha \stackrel{}{\sim} \text{Beta}(1,\alpha),~~i = \{1,2,...\}\nonumber\\
    \pi_i = \nu_i \prod_{\ell=1}^{i-1}(1-\nu_\ell),~~i=\{1,2,...\}.
\end{gather}
This process can be understood by imagining a unit probability stick being continually partitioned, with the proportion of the remaining stick to be broken off chosen according to a beta distribution parameterized by $(1,\alpha)$.

A draw from the DP (i.e., $G \stackrel{}{\sim} DP(\alpha,G_0)$) is a discrete, infinite random object that can be expressed by

\begin{equation}
    \label{eq:DPdraw}
    G = \sum_{i=1}^{\infty} \pi_i\delta_{\theta_i},~~i=\{1,2,...\},
\end{equation}
\noindent
where $\theta_i$ is the $i^{th}$ of the countably infinite atoms drawn i.i.d. from a base distribution, $G_0$. That is,
\begin{equation}
    \label{eq:atoms}
    \theta_i\lvert G_0 \stackrel{i.i.d.}{\sim} G_0.
\end{equation}

In our case, $G_0$ is typically the joint conjugate prior for the means and covariances which specify the Gaussian components (i.e., the normal-inverse-Wishart distribution, equation \ref{eq:NIW})~\cite{Gorur}. In other words, the atoms of the DP parameterize Gaussians centered around the base hyperparameters. The concentration hyperparameter, $\alpha$, determines the extent to which the atoms cluster around $G_0$. Marginalization over the infinite sequence of mixture proportions in the so-called Chinese Restaurant Process (see \cite{Fox}) exposes the \textit{preferential attachment} of the cluster assignments. This is integral to instantiating as few components as necessary given the observed data. Hierarchical models involving hyperpriors over the hyperparameters of $G_0$ can be constructed to guard against poor model initializations, however we restrict our attention to the simpler case of fixed hyperparameters.


To compute the postdictive surprise, we require the predictive density, given by

\begin{equation}
    \label{eq:pred1}
    p(x_{N+1}\lvert x_1,...,x_N,\alpha,G_0) = \int p(x \lvert \theta) p(\theta\lvert x_1,...,x_N,\alpha,G_0)d\theta.
\end{equation}

However, the DP prior precludes an analytic closed form for the posterior distribution, $p(\theta\lvert x_{1:N},\alpha,G_0)$. Although MCMC methods are a common method for approximating such densities, inference of model parameters by sampling methods are typically slow, and scale poorly as the number of parameters or data points increases \cite{MCMC}.
Additionally, convergence metrics are heuristic at best. In contrast, variational methods select a simpler family of distributions whose posterior density is ideally able to approximate the true posterior by optimizing a set of variational parameters. These parameters are optimized with respect to the evidence lower bound (ELBO), a constraint on the log marginal likelihood of the data, which is straight-forwardly related to the divergence between the variational posterior and the true posterior. Thus, convergence -- at least to a local optimum -- is well-defined. Variational inference methods for Dirichlet Process mixture models were first introduced in \cite{blei}, and in this paper we make use of the scikit-learn implementation \cite{sklearn}, available as of version 0.18.

The variational approximation is proposed to take the following form:

\begin{equation}
    \label{eq:q}
    q(\mathbf{\nu},\mathbf{\theta},\mathbf{z}) = \prod_{k=1}^{K-1} q_{\gamma_k}(\nu_k) \prod_{k=1}^K q_{\tau_k}(\theta_k) \prod_{n=1}^N q_{\phi_n}(z_n)
\end{equation}
\noindent
Here, $\{\mathbf{\gamma},\mathbf{\tau},\mathbf{\phi}\}$ are the variational parameters subject to coordinate ascent optimization. $q_{\gamma}$ are beta distributions parameterized by the individual stick lengths, $\nu_k$. $q_{\tau}$ are in our case Gaussians parameterized by $\theta_k = \{\mu_k,\Sigma_k\}$, although extension to general exponential families is possible. $q_{\phi_n}$ are multinomial, parameterized by indicator variables $z_n$, which denote the component to which the observation $x_n$ is assigned. To speed up inference, a truncation on the maximum number of possible states is imposed on the variational approximation, similar to truncation in methods such as blocked Gibbs sampling~\cite{blei}. This value, $K$, is itself a variational parameter which can be fixed or optimized with respect to the ELBO. $K$ was fixed in our work to 30 unique components. Under this approximation, the resulting posterior predictive distribution needed for computing postdictive surprise can be neatly factored as expectations with respect to the variational distribution:

\begin{equation}
    \label{eq:predictive_factorized}
    p(x_{N+1}\lvert x_1,...,x_N,\alpha,G_0) \approx \sum_{k=1}^{K}\mathbb{E}_q\big[\pi_k \big]\mathbb{E}_q\big[p(x_{N+1}\lvert\theta_k)\big]
\end{equation}

For many machine learning algorithms, decay in the postdictive surprise might be sufficient to demarcate useful data from superfluous data during training. However, it is often the case that temporal relationships between appliance states are learned and contribute to inference. Such methods would include Hidden Markov Models (HMMs) and their many extensions, more recent deep learning techniques such as those based on Recurrent Neural Networks, and many more. In the interest of simplicity, we restrict the notion of ``transitional surprise'' to the Markovian sense. That is, we treat the state sequence as a Markov chain, such that the current state of the system is determined only by the state before it. For a system of $K$ appliance states, this transitional surprise constitutes comparing the rows of the $K\times K$ transition matrix. This approximation to the dynamics is clearly crude, but even weak convergence of the transition matrix to some stationary form can prove useful.

To summarize, for each sliding window of $w$ events, preceded by $N$ events, we compute the (approximate) postdictive surprise as:

\begin{equation}
    \label{eq:post_surprise}
    S_{o} = d\big[p(x_{N+1}\lvert x_{1:N},\alpha,G_0)~\lvert \lvert ~p(x_{N+w+1}\lvert x_{1:(N+w)},\alpha^{*},G_0)\big],
\end{equation}

\noindent
where $d$ is some divergence metric (usually Kullback-Leibler divergence), and $\alpha^*$ is the posterior update for the concentration parameter if a prior was placed on it.

Over the same window of $w$ events, we compute the transitional surprise over the truncated maximum number of states $K$ as:

\begin{equation}
    \label{eq:trans_surprise}
    S_t = \sum_{j=0}^{w}\sum_{k=1}^{K} d\big[T_k(z_{1:N+i})\lvert\lvert T_k(z_{1:N+i+1})\big],
\end{equation}
\noindent
where at time $t$, $T_{j,k} = p(z_{t+1}=k \lvert z_t=j)$. The notation $T_k(z_{1:N+i})$ denotes the transition row built using event indicators $z$ for observations $1,2,...,N+i$.

In order to simplify the concept of a surprise threshold under which data is no longer considered surprising, $S_o$ and $S_t$ are normalized according to their maximum values. Since the initial value of the above divergences can certainly be exceeded as observations are made, the maxima were updated and preceding surprise values were renormalized to the revised maxima. Since in an online setting it would be unreasonable to wait indefinitely for surprising windows, we suggest a patience parameter, $\rho$. In the experiments that follow, we used $\rho=100$; that is, 100 windows are observed beyond the most recent window exceeding the surprise threshold. If no other windows exceed the threshold, the previously surprising window is returned as the cutoff point.

\section{Experiments}\label{sec:exp}
To explore the usefulness of a surprise threshold, we made use of NILMTK, an open-source toolkit developed for NILM research \cite{batra2014nilmtk, batra2019towards}. NILMTK includes implementations of some benchmark algorithms including traditional

\begin{enumerate}

    \item \emph{Denoising Autoencoders (DAE)}: treat load disaggregation as noise reduction problem, in which the aggregate signal is seen as noisy version of an appliance signal. This special kind of neural network is typically implemented following a symmetrical architecture has originally been introduced to perform representation learning \cite{bonfigli2018denoising}.

    \item \emph{Recurrent Neural Networks (RNN)} have been successfully applied to a variety of time series problems. For NILM, RNNs have been proposed in \cite{kelly15neuralnilm}, where the nets were trained to detect signatures of appliance within smart meter data. In this work, the RNN architecture proposed by \cite{krystalakos18windowgru} is being used, which incorporates Long short-term memory (LSTM) cells.

    \item \emph{Sequence-to-Sequence Optimization (Seq2Seq)} is a technique using neural networks, introduced in \cite{zhang2018sequence}. The basic idea of this approach is to learn the mapping between the aggreagte input window and the output window, which is a sequence of power consumption values associated with a certain appliance.

    \item The \emph{Sequence-to-Point Optimization (Seq2Point)} technique builds on neural networks and is closely related to \ac{S2S}. The main difference between these two techniques lies in the output layer of the architecture, where S2P was designed to forward the midpoint of the output window.~\cite{zhang2018sequence}

    \item The \emph{Window GRU} architecture, introduced in \cite{krystalakos18windowgru}, relies on Gated Recurrent Units (GRU). Compared to architectures based on LSTMs, this architecture is simpler, integrates fewer neurons per layer and therefore, was shown to be more computationally efficient while having lower memory demand.

\end{enumerate}


To establish a relationship between algorithm performance and the proposed surprise metrics, three houses from the REFIT dataset~\cite{refit} were selected for study using the above disaggregation methods. The included appliances in these experiments were the dish washer, the washing machine, the refrigerator, the kettle, and the toaster. The Mean-Absolute Error (MAE) was used as a performance metric, defined by
\begin{gather}
    MAE = \frac{1}{N}\sum_{t=1}^{N}\lvert \hat{x}_t  - x_t \rvert,
\end{gather}
\noindent
where $N$ is the number of samples, and $\hat{x}_t$ is the predicted load at sample $t$. For each house, the available data was split into a training set and test set by a 90\%/10\% split. 15\% of the training set was reserved for validation. The surprise metric was computed on the remaining training data, such that each algorithm was training and validating on the same data. Each algorithm was trained over 15 epochs using Adam optimization with a batch size of 1024 samples. For a given house, each algorithm had its random seed fixed across surprise-based training set reductions, removing initialization variability from their appliance-averaged performance. Preprocessing of the data such as normalization was handled internally by NILMTK.

\begin{figure*}[ht]
\centering
    \subfloat{\centering
      \input{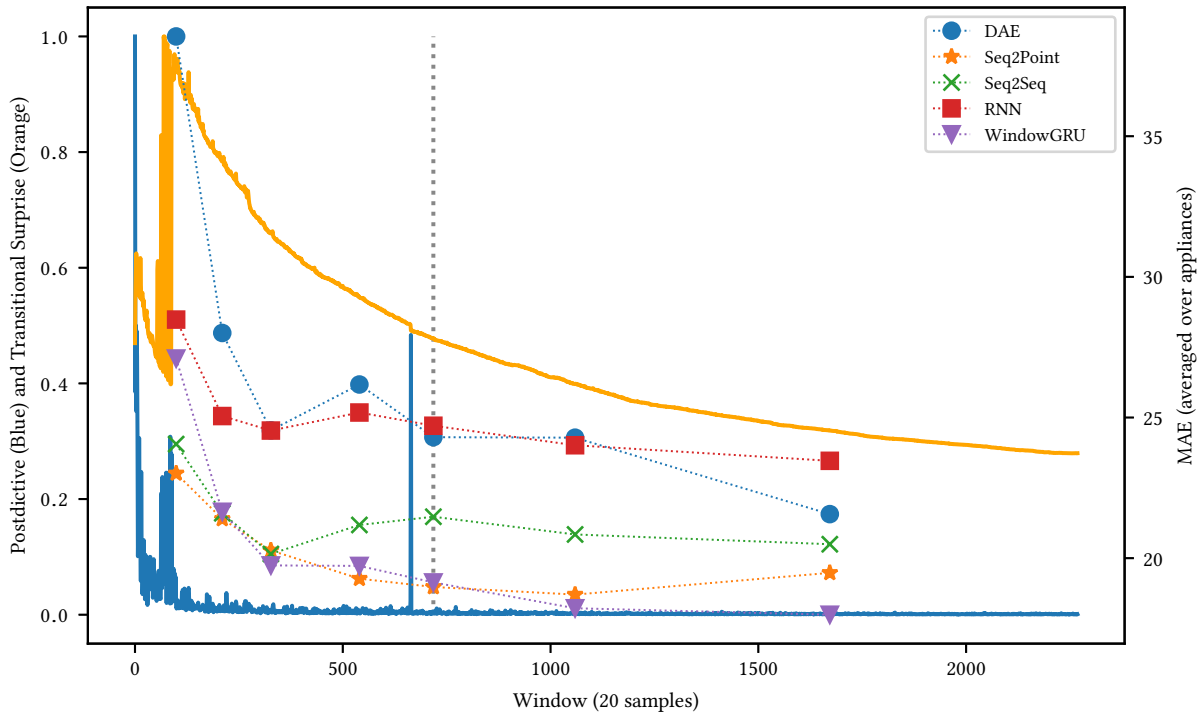}\label{fig:h2_mae:input}
    }
    \caption{Appliance-averaged MAE performance, REFIT House 2}
    \label{fig:h2_mae}
\end{figure*}

\begin{figure*}[ht]
\centering
    \subfloat{\centering
      \input{refith3_mae.pgf}\label{fig:h3_mae:input}
    }
    \caption{Appliance-averaged MAE performance, REFIT House 3}
    \label{fig:h3_mae}
\end{figure*}

\begin{figure*}[ht]
\centering
    \subfloat{\centering
      \input{refith5_mae.pgf}\label{fig:h5_mae:input}
    }
    \caption{Appliance-averaged MAE performance, REFIT House 5}
    \label{fig:h5_mae}
\end{figure*}


Figures \ref{fig:h2_mae}, \ref{fig:h3_mae}, and \ref{fig:h5_mae} show the behaviour of the MAE for the average appliance across the benchmark methods for houses 2, 3, and 5, respectively. The postdictive and transitional surprise was computed using Jensen-Shannon divergence, defined between two distributions $p$ and $q$ by:

\begin{equation}
    \label{eq:JSdiv}
    d_{JS}(p\lvert\lvert q) = \frac{d_{KL}(p\lvert\lvert m)d_{KL}(q\lvert\lvert m)}{2},
\end{equation}
\noindent
where $m$ is the point-wise mean of $p$ and $q$, and $d_{KL}$ is the Kullback-Leibler divergence, given by
\begin{equation}
\label{eq:KLdiv}
	d_{KL}(p\lvert\lvert q) = \sum_{x \in X}^{}p(x) \cdot log\bigg(\frac{p(x)}{q(x)}\bigg).
\end{equation}

Given the max-value normalization, the postdictive and transitional surprise values can be interpreted as the fraction of the maximum observed surprise, rather than the value of the JS-divergence itself.

Although of course no sharp transition exists between an optimally and sub-optimally sized training set, the behaviour of these algorithms' MAE in the three REFIT houses suggest that performance can indeed stagnate. Additional similar data, especially in houses 2 and 3, seem unlikely to appreciably improve performance. An example surprise threshold is shown in figures \ref{fig:h2_mae}, \ref{fig:h3_mae}, and \ref{fig:h5_mae} as a dotted grey line, indicating an approximate point where performance began to plateau. This cutoff was chosen as a joint threshold over postdictive and transitional surprise, defined by:
\begin{equation}
    \label{eq:cutoff}
    S_o(w:w+\rho) \leq 0.01~\&~~S_t(w:w+\rho)~\leq 0.05,
\end{equation}
\noindent
where again, $w$ is the window size and $\rho$ is the patience parameter. We used this threshold for further study regarding the potential regularizing effect of surprise-based training cutoff.

In \cite{murray}, disaggregation performance on unseen homes in the same dataset as well as different datasets were examined. By their choice of architectures, the authors restricted the number of tunable parameters relative to the existing literature. They also made use of early stoppage with an aggressive patience parameter to terminate training. With these complexity and temporal regularization methods, they showed intra- and inter-dataset transferability with minimal performance losses relative to their chosen baseline. Nevertheless, these methods still make use of all available training data. Bayesian surprise metrics provide an attractive alternative/supplement to early stoppage, which by contrast truncate the training set entirely. We examined the MAE performance of each algorithm when trained on the full REFIT house 3 and the surprise-based subset determined by the joint threshold in equation \ref{eq:cutoff}. Table \ref{tbl:cross} shows the appliance-averaged MAE performance of each benchmark method when tested on REFIT house 5. All but one method showed improved cross-house transferability with a restricted training set, giving some substance to the claim that truncating the training set may provide regularization against overfitting.

\begin{table}[]
\caption{REFIT Cross-house ($3 \rightarrow 5$) MAE for full and cutoff training}\label{tbl:cross}
\centering
\begin{tabular}{c|cc}
\toprule
 Benchmark Method & Full Training & Cutoff Training\\\midrule
WindowGRU & 37.83 & \textbf{33.03} \textcolor{green}{$\downarrow$} \\
DAE & 34.78 & \textbf{33.00} \textcolor{green}{$\downarrow$}\\
RNN & 32.54 & \textbf{30.62} \textcolor{green}{$\downarrow$}\\
Seq2Seq & \textbf{27.17} & 29.43 \textcolor{red}{$\uparrow$}\\
Seq2Point & 26.85 & \textbf{26.74} \textcolor{green}{$\downarrow$}\\
\bottomrule
\end{tabular}
\end{table}

Finally, to illustrate the usefulness of including the concept of transitional surprise, we explored the performance of a popular super-state Hidden Markov Model~\cite{makonin_sshmm}. Clearly, a Markovian model should suffice to show whether our Markovian notion of transitional surprise is useful. We used house 1 from the Rainforest Automation Energy (RAE) dataset~\cite{makonin2018rae}, which consists of two blocks: a 9 day block beginning on February 7, 2016, and a 63 day block beginning March 6, 2016. Block 1 was used as the test set, and block 2 (and its surprise-based subset) was used for training the models. The seven appliances used for training were the clothes washer and dryer, refrigerator, dish washer, furnace/hot water unit, and the heat pump.

\begin{figure*}[ht]
\centering
    \subfloat{\centering
      \input{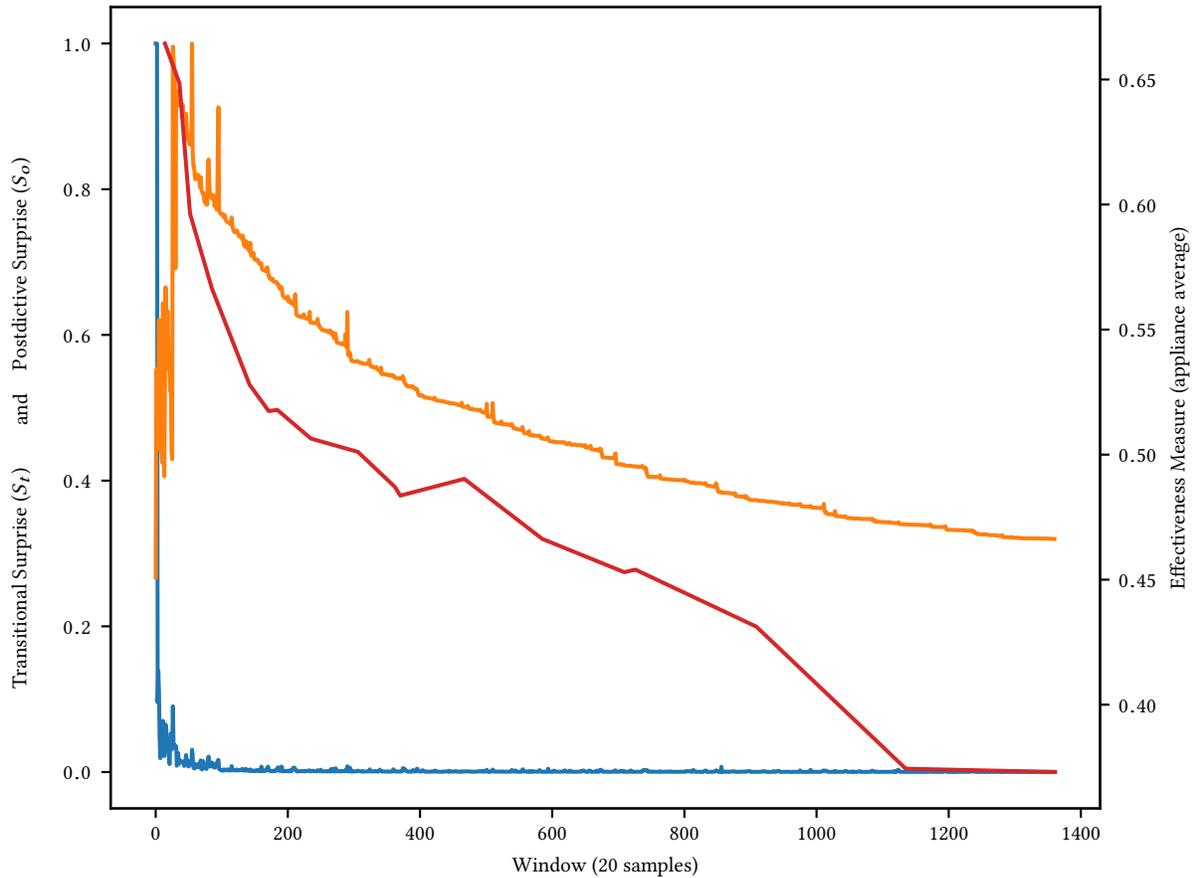}\label{fig:sshmm}
    }
    \caption{Effectiveness measure averaged over 7 appliances as a function of surprise-based training cutoff}
    \label{fig:sshmm}
\end{figure*}

Figure \ref{fig:sshmm} shows the Van Rijsbergen's effectiveness measure (defined simply as $1-$~F1-score) as a function of cutoff point during training. This measure decays slightly faster than that of the transitional surprise, but significantly after the postdictive surprise had converged. This lends credence to the claim that postdictive surprise is an unreliable metric for terminating training in the general case. The difference in decay rate between transitional surprise and the effectiveness measure is understandable given that the SSHMM by definition encodes the Markovian dynamics between super-states of the user's home. The super-state of the home at a given instant in time can be thought of as the complete description of the home, denoting the operational mode of each appliance in the house. Each instant in time increments the underlying transition distributions between super-states of the home, rather than individual appliance states. This will in general encode the state dynamics more efficiently since there is more information used per time-step. Nevertheless, the basic notion of transitional surprise introduced here allows a useful overestimate of the learning rate of the system dynamics. Notably, the behaviour of the effectiveness measure in this case calls into question the specific values given for the joint threshold in equation \ref{eq:cutoff}. Here, a threshold on the transitional surprise of $\approx 0.4$ seems adequate to predict stagnant performance improvements for this dataset. Significant exploration with all available datasets is needed to further narrow down acceptable threshold values.


\section{Conclusions} \label{sec:conclusions}

Ultimately, the concept of surprise involves comparison over distributions as they are updated given new observations. The most useful such distributions are unavoidably model-specific. For example, surprise could be defined relative to the latent space in methods such as the DAE, or it could be defined relative to nonlinear auto-regressive dependencies in more complex graphical models. Nevertheless, there are features intrinsic to the data itself that could be used to predict the usefulness of more data in a model-agnostic way. This work explored a postdictive surprise defined over the likelihood of a non-parametric GMM. The mixture model was updated with windows of events defined by first-differences in the block-filtered raw signal exceeding a pre-specified threshold. Furthermore, we explored a transitional surprise defined in a Markovian sense, which was described by the transitional relationships between latent states as determined by the state assignments of the GMM. This crude approximation to the system dynamics was shown to be useful relative to a strictly postdictive notion of surprise, at least in an HMM-based application. An approximate joint threshold was determined by examining the MAE performance of five benchmark methods supported by NILMTK over three REFIT homes. This threshold was used to explore the potential regularizing effect of a surprise-based training cutoff. This is similar to the use of early-stoppage, which is a common method to protect against over-fitting and aid in the transferability of learned parameters. Relative to training over the full REFIT house 3, training on the surprise-based subset showed improved MAE for all but one method when testing over REFIT house 5. This supports the claim that Bayesian surprise can be a useful metric in predicting over-fitting and potentially improve generalization to unseen houses or datasets.

Further experiments may show that convergence of transitional and postdictive surprise are only weakly indicative of a plateau in model performance, and that models continue to improve when using additional, repetitive data. In this case, it is unlikely that researchers would make use of a surprise-based cutoff in their final training of a particular model. However, during development it may be highly desirable to merely gauge the effectiveness of new methods or network modifications without spending copious amounts of time retraining using all available data. In these cases, truncating the training set using surprise-based methods allows a significant reduction in research costs, both in terms of computational time spent training and research time spent trying to optimize what may prove to be fruitless methods.

Moreover, postdictive surprise using non-parametric mixture models naturally extends to online settings, where deployed NILM algorithms quickly become obsolete without the flexibility to adapt to new appliances or appliance replacements.

Lastly, this work suggests a general rule of \textit{diversity over quantity} of data. This may help inform the development of future datasets, improving time-to-publication for dataset producers as well as expediting dataset availability for the research community as a whole.

An important extension of the current work is to explore cross-dataset performance. Similarities between the two REFIT houses in table \ref{tbl:cross} is likely unrepresentative of the general use-case for NILM. Also left to future work is to explore alternative models for transitional surprise such as constructing super-states from the observed appliance modes. Additionally, future work may include sub-modelling for each component observed in the non-parametric mixture model. This would permit modelling multiple appliance modes in the same range of power values, where Bayesian surprise could further be computed over the sub-model parameters. This extension would be highly valuable to an online setting to track new appliance mode activations.

\balance
\bibliographystyle{ACM-Reference-Format}
\bibliography{sample-base}

\end{document}